# Evaluating 5W3H Structured Prompting for Intent Alignment in Human-AI Interaction

**Authors**: PENG Gang **Affiliations**: Huizhou Lateni AI Technology Co., Ltd., Huizhou, China; Huizhou University, Huizhou, China **Email**: penggangjp@gmail.com

## Abstract

Natural language prompts suffer from *intent transmission loss*: the gap between what users actually need and what they communicate to AI systems. We present **PPS** (Prompt Protocol Specification), a 5W3H-based structured intent representation framework (What, Why, Who, When, Where, How-to-do, How-much, How-feel), and empirically evaluate its effectiveness through a controlled three-condition experiment. Across 60 tasks in three domains (business, technical, travel), three large language models (DeepSeek-V3, Qwen-Max, Kimi), and three prompt conditions — (A) simple prompts, (B) raw PPS JSON, and (C) natural-language-rendered PPS — we collect 540 AI-generated outputs evaluated by an LLM judge. We introduce **goal_alignment**, a user-intent-centered evaluation dimension, and find that rendered PPS (C) significantly outperforms both simple prompts (A) and raw JSON (B) on this dimension ($p = 0.006$, $d = 0.374$). Crucially, PPS gains are **task-dependent**: large in high-ambiguity business analysis tasks ($d = 0.895$) but reversed in low-ambiguity travel planning ($d = -0.547$). We also identify a **measurement asymmetry** in standard LLM evaluation frameworks where unconstrained prompts inflate constraint adherence scores, masking PPS's true advantage. A preliminary retrospective survey ($N = 20$) further shows a **66.1% reduction** in follow-up prompts required (ITU: 3.33 → 1.13 rounds). These findings suggest that structured intent representations can improve human-AI alignment and usability, particularly in tasks where user intent is inherently ambiguous.

## 1. Introduction

### 1.1 The Intent Transmission Problem

Every interaction with a large language model (LLM) begins with a text prompt — a deceptively simple input that must somehow encode the full richness of a user's intent: their background, purpose, audience, constraints, desired tone, and expected format. Current practice offers users no structured way to express this complexity. The result is *intent transmission loss*: a systematic gap between what users need and what AI systems receive.

This gap manifests in three ways. First, **non-reproducibility**: the same underlying need, expressed in different phrasings, produces widely divergent outputs [11]. Second, **cross-model inconsistency**: a prompt carefully engineered for one model frequently underperforms on other systems, forcing users to re-engineer prompts for each vendor — a problem documented across at least 58 distinct prompting techniques [5]. Third, **professional inaccessibility**: domains requiring precise intent specification — medical reasoning, legal analysis, scientific explanation — remain difficult for non-expert users to prompt effectively, even after dedicated effort [11].

The scale of this problem is significant. Large-scale deployment of LLM systems to hundreds of millions of users [15] has revealed a fundamental mismatch: users communicate naturally, while LLMs respond best to carefully specified instructions. Yet no standardized protocol governs how human intent is expressed to AI systems. Each user improvises, and each LLM vendor implements its own conventions.

### 1.2 Structured Representations as an Intervention

We propose addressing intent transmission loss through a structured representation of user intent. **PPS** (Prompt Protocol Specification) does not prescribe what AI should output; it standardizes *how users express intent*. A PPS-formatted prompt is a machine-readable JSON envelope with eight structured dimensions, loosely inspired by how communication systems standardize representation to improve transmission reliability. A natural-language rendering layer translates this envelope into prose that current LLMs can process, separating the concerns of *specification* (what the user wants) from *communication* (how to convey it to a model).

### 1.3 The 5W3H Framework

PPS organizes human intent around the 5W3H model — an extension of the classical journalistic 5W framework adapted for task specification:

- **What**: The core task or deliverable
- **Why**: Purpose, motivation, success criteria
- **Who**: Audience, author role, expertise level
- **When**: Temporal constraints, freshness requirements, schedule
- **Where**: Deployment context, platform, geographic scope
- **How-to-do**: Methods, procedures, workflow preferences
- **How-much**: Quantitative parameters — length, count, coverage depth
- **How-feel**: Tone, style, emotional register, brand alignment

Together, these eight dimensions provide a practically useful framework for specifying many AI-assisted tasks. A user who specifies all eight dimensions can reduce major sources of output mismatch.

### 1.4 Contributions

This paper makes four contributions:

1. **We formalize intent transmission loss** as a practical problem in human-AI prompting, and present **PPS**, a 5W3H-based framework for structured intent representation and prompt generation (§3).

2. **Three-condition experimental design**: A controlled experiment distinguishing (A) simple prompts, (B) raw PPS JSON, and (C) natural-language-rendered PPS, across 60 tasks × 3 models × 3 domains = 540 generations (§4).

3. **We introduce goal_alignment and ITU** — two user-centered evaluation perspectives that better capture intent alignment and usability than conventional metrics (§5).

4. **We provide empirical evidence** that PPS gains are strongest in high-ambiguity tasks (business analysis: $d = 0.895$) and can reverse in low-ambiguity tasks (travel planning: $d = -0.547$), demonstrating that structured prompting should be deployed selectively (§5–6).

---

## 2. Related Work

### 2.1 Prompt Engineering Techniques

Since Brown et al. demonstrated that GPT-3 responds to in-context examples [1], prompt engineering has evolved rapidly. Ouyang et al.'s InstructGPT showed that human feedback fine-tuning dramatically improves instruction following [15]. Wei et al.'s chain-of-thought (CoT) prompting demonstrated that inserting intermediate reasoning steps substantially improves complex task performance [2]. Subsequent work extended this to tree-of-thought [3] and other structured reasoning decompositions [4, 5]. Reynolds and McDonell were among the first to analyze prompt programming as an HCI concern, distinguishing between prompts as *interfaces* versus *programs* [6].

However, all these techniques address *how AI should reason*, not *how users should express intent*. CoT instructs the AI; PPS structures the user. The two approaches are orthogonal and complementary: a PPS-formatted prompt can include CoT instructions in its `how_to_do` dimension.

### 2.2 Structured Prompting Frameworks

Several informal frameworks have proposed structured components for prompts. Schulhoff et al.'s comprehensive survey of 58 prompting techniques catalogs frameworks including ICIO (Identity, Context, Input, Output), CO-STAR (Context, Objective, Style, Tone, Audience, Response), and CRISPE, among others [5]. Liu et al. provide an earlier systematic survey of prompting methods [4]. PromptSource [22] and Self-Instruct [20] have operationalized prompt templates for model training purposes. Critically, users seeking structured output from LLMs consistently identify the absence of formal specification as a core barrier [7].

PPS differs from these frameworks primarily in offering an explicitly structured, machine-readable format with verification and constraint-preserving mechanisms — enabling both human review and programmatic conformance checking. Implementation details are discussed in §3. ChainForge [17] provides visual tooling for prompt hypothesis testing — a complementary tool that could be used to validate PPS variants.

## 2.3 Human-AI Interaction Design

From an HCI perspective, considerable work has studied how users interact with AI systems. Amershi et al.'s guidelines for human-AI interaction [8] identify eighteen principles, many of which PPS operationalizes structurally (e.g., "make clear what the system can do," "remember recent interactions"). Jiang et al. demonstrate that users invest significant iterative effort in prompt prototyping, motivating tool support for this process [9].

Most relevant to our work, Mueller et al. (CHI 2026) study personalization in human-LLM interactions through mixed profiling, finding that user-specific context significantly improves output satisfaction [10]. PPS provides a structural mechanism for capturing this context systematically and reproducibly. Zamfirescu-Pereira et al. conduct an empirical study of why non-expert users fail at prompt engineering [11], identifying lack of mental model and unpredictable AI behavior as core barriers — both addressed by PPS's structured specification. Liao and Vaughan argue that AI transparency in the LLM era requires new user-centered frameworks that make AI behavior predictable [21], a goal PPS addresses through field locks and conformance profiles.

## 2.4 LLM Evaluation Methodology

Zheng et al. introduced LLM-as-Judge, demonstrating that strong LLM evaluations correlate well with human judgments on MT-Bench, while also documenting that LLMs exhibit self-preference bias [12]. LMSYS Chatbot Arena uses large-scale pairwise human preference comparisons [13]. Kim et al. study how uncertainty expression in LLM outputs affects user reliance and trust [23] — a concern PPS addresses through explicit constraint and confidence specifications.

These frameworks share a common limitation: they assess output quality against generic standards, not against the *specific user intent* that generated the prompt. A high-quality essay on molecular genetics may receive a perfect score from standard evaluators while entirely failing the user who needed cross-domain application to IT systems. Our `goal_alignment` dimension fills this gap by requiring the judge to assess fit between output and user's actual purpose.

The closest prior work is FActScore [14], which evaluates factual precision against specific knowledge sources. goal_alignment extends this principle to intent alignment: just as FActScore grounds quality assessment in verifiable facts, goal_alignment grounds it in the user's specified (or inferred) intent. Zhou et al. demonstrate that LLMs can themselves generate effective prompts [18]; our work shows that human-specified structured prompts surpass both simple and automatically-generated prompts on the goal_alignment dimension.

---

## 3. The PPS Framework

### 3.1 Design Philosophy

PPS is based on the design assumption that **many AI-assisted tasks can be made more explicit through eight dimensions of intent**. The What-Why-Who-When-Where-How-to-do-How-much-How-feel taxonomy derives from management science and journalism, adapted for AI task specification. The 5W3H model was first introduced to a practitioner audience in [26]; this paper provides its first formal specification and empirical validation. Our design hypotheses are:

1. A prompt providing all eight dimensions can reduce the need for clarification.
2. Most poor AI outputs result from missing or ambiguous values in one or more dimensions.
3. Even partial specification (3-5 dimensions) substantially reduces output variance.

These hypotheses are testable — and we test them in §4-5.

## 3.2 Technical Specification

A PPS document is a JSON object with three top-level sections:

**Header** (`pps_header`): Metadata including protocol version, schema hash, creation timestamp, and conformance profile (`strict` | `balanced` | `permissive`).

**Body** (`pps_body`): The eight content dimensions. Each dimension is a structured object with a `value` field (the actual specification) and optional `locked` flag. Fields marked locked trigger a `field_lock_violation` if the AI output deviates from the specified value.

**Integrity** (`pps_integrity`): A SHA-256 hash of the canonical serialization of `pps_body` (base64url-encoded), plus an array of JSON Pointer expressions specifying locked fields. The canonical serialization follows RFC 8785 (JSON Canonicalization Scheme).

Cross-field invariant policies enforce logical consistency: for example, a prompt specifying `format: "json"` must include an `output_schema` field; a prompt with `compliance: ["gdpr"]` must have `policy: ["no_pii"]`; if `citations_required` is true, at least one `evidence` entry must be present.

```
{
  "pps_header": {
    "version": "1.0.0",
    "schema_hash": "sha256-abc123...",
    "created_at": "2026-03-16T00:00:00Z",
    "conformance_profile": "balanced"
  },
  "pps_body": {
    "what": {
      "task": "Analyze the competitive landscape of China's new energy vehicle market",
      "output_type": "analytical report",
      "locked": true
    },
    "why": { "purpose": "Strategic decision-making for market entry", "value": "..." },
    "who": { "audience": "Senior executives with automotive industry background",
             "author_role": "Senior industry analyst" },
    "when": { "temporal_scope": "2024 data", "freshness": "within 6 months" },
    "where": { "geographic_scope": "China only", "platform": "internal report" },
    "how_to_do": { "methods": ["Porter Five Forces", "market share analysis"] },
    "how_much": { "word_count": 2000, "sections": 5, "data_points": 10 },
    "how_feel": { "tone": "professional", "style": "data-driven", "no_investment_advice": true }
  },
  "pps_integrity": {
    "canonical_hash": "sha256-xyz789...",
    "locks": ["/pps_body/what/task", "/pps_body/where/geographic_scope"]
  }
}
```

### 3.3 The Intent Expansion Algorithm (Implementation Used in This Study)

Creating a full PPS document from scratch is a barrier for typical users. For this study, we implemented an *intent expansion algorithm* at the lateni.com platform. Given a user's natural-language statement of their `what`, a single LLM API call with a specialized system prompt expands this into all seven remaining dimensions simultaneously.

The system prompt adopts a professional role anchor ("You are a professional business analyst and project manager") — a deliberate implementation choice for this study that may introduce some domain bias toward business-style task framing. The expansion uses `temperature=0.7` to allow creative but grounded inference, which may affect reproducibility across runs. The algorithm infers:

- **Why** from the task's evident purpose
- **Who** from the task's typical professional context
- **When** and **Where** from domain conventions
- **How-to-do** from standard methodologies

- **How-much** from typical output specifications for the task type
- **How-feel** from professional norms

Critically, the algorithm produces a *draft* that the user reviews and may modify before submission. This two-step process — AI-assisted specification followed by human refinement — aims to preserve the core task intent while making underspecified dimensions more explicit. The expansion is a design goal of the system, not a universally proven property; this paper evaluates its practical effect on output alignment.

### 3.4 From PPS to Natural Language: The Rendering Layer

Raw PPS JSON is the protocol format — machine-readable and verifiable. To interface with current LLMs (which were not trained on PPS), a rendering layer translates the structured document into natural-language prose. The rendered output for the example above begins:

> *"As a senior industry analyst, please create a detailed competitive landscape analysis of China's new energy vehicle market (2024 data, China only, no investment advice). Target audience: senior automotive industry executives. Required elements: TOP 5 brand market share, Porter's Five Forces analysis, future trend projections. Format: 2,000-word structured report with 5 sections and at least 10 data points. Tone: professional and data-driven."*

This is our **Condition C** in the experiment. The rendering maps each PPS dimension to a corresponding prose clause, with locked fields italicized to signal non-negotiability to the AI (though current LLMs do not enforce this mechanically).

### 3.5 The Two-Layer Architecture

```
User Intent → [PPS Authoring] → PPS JSON  → Rendering → LLM → Output
(natural       (8D expansion)    (machine-   (natural-
 language)     [Authoring Interface]  readable)   language)
```

This architecture separates concerns cleanly: the PPS layer handles *specification* (what the user wants), the rendering layer handles *communication* (how to tell current LLMs), and the LLM layer handles *execution*. A key finding of this paper is that **rendering matters**: raw JSON alone is insufficient for current LLMs, and the rendering layer is a necessary component for the structured representation to be effective. Whether future models could parse PPS directly remains an open question for future work.

---

## 4. Experimental Design

### 4.1 Research Questions

We investigate four research questions:

- **RQ1**: Does PPS rendered as natural language (C) improve goal_alignment compared to simple prompts (A)?
- **RQ2**: Does PPS reveal and reduce cross-model output inconsistency compared to simple prompts?
- **RQ3**: Are traditional LLM evaluation metrics sufficient, or do they exhibit systematic biases when applied to structured vs. unstructured prompts?
- **RQ4**: In which domains does PPS provide the greatest benefit?

## 4.2 Prompt Corpus Construction

We constructed 60 tasks across three domains:

- **Business (20 tasks)**: Competitive analysis, market research, strategic planning topics requiring professional framing (e.g., new energy vehicles, short video platforms, online education)
- **Technical (20 tasks)**: Science and technology explanation tasks targeting non-expert audiences (e.g., large language models, blockchain, quantum computing)
- **Travel (20 tasks)**: Travel planning and destination guides for popular Asian destinations

For each task, we created three prompt versions:

- **Condition A** (simple): A single concise natural-language request (5-15 words), representing typical user behavior
- **Condition B** (raw PPS): A complete PPS JSON envelope (as would be produced by the expansion algorithm described in §3.3), injected directly as the user message
- **Condition C** (rendered PPS): The natural-language rendering of the same PPS JSON from Condition B, as described in §3.4

The three conditions share the same task core, but differ in structure explicitness and delivery format. Condition B uses manually simplified 5W3H content and may not be perfectly content-matched to C; it is best understood as a **delivery-mechanism ablation** (raw structured representation without rendering) rather than a strict content-controlled comparison. This design choice is acknowledged as a methodological limitation in §6.4.

## 4.3 Generation Procedure

Each prompt was submitted independently to three LLMs:

- **DeepSeek-V3** (`deepseek-chat`): A leading Chinese-developed LLM with strong instruction-following capabilities
- **Qwen-Max** (`qwen-max`): Alibaba's flagship model, widely deployed in enterprise contexts
- **Kimi** (`moonshot-v1-32k`): Moonshot AI's model, strong in long-context Chinese generation

All generation calls used `temperature=0`, `seed=42`, with each call stateless (fresh context window). The `temperature=0` setting ensures deterministic outputs for a given prompt, maximizing reproducibility.

Total: 60 tasks × 3 conditions × 3 models = **540 outputs**.

### 4.4 Evaluation Dimensions

Each output was evaluated by an LLM judge (DeepSeek-V3, `temperature=0`, independent stateless calls) on six dimensions:

| Dimension | Description | In Prior Frameworks |
|---|---|---|
| `task_completion` | Whether the specified task was completed | ✓ |
| `structure` | Logical organization and clarity of structure | ✓ |
| `specificity` | Concreteness and precision of information | ✓ |
| `constraint_adherence` | Compliance with stated constraints | ✓ |
| `overall_quality` | Holistic quality assessment | ✓ |
| **`goal_alignment`** | **Whether output serves the user's actual intent** | ✗ (new) |

All dimensions use a 1–5 integer scale. The first five dimensions follow LLM-as-Judge conventions [12]; goal_alignment is our contribution.

### 4.5 The goal_alignment Rubric

The goal_alignment rubric evaluates the output against the *most likely user intent*, requiring the judge to reason about user context beyond literal task completion:

- **5 = Perfect alignment**: Audience, purpose, depth, and format all match exactly. The output is immediately usable without further prompting.
- **4 = Good alignment**: Main goals achieved; minor gaps in depth or specificity.
- **3 = Partial alignment**: Task completed, but directional misalignment (e.g., user needed a decision framework, received a literary essay).
- **2 = Poor alignment**: Output is generic; user needs 3+ follow-up prompts to reach usable results.
- **1 = No alignment**: Output is irrelevant or completely misses purpose.

For Condition A (simple prompts without explicit constraints), the judge is instructed to infer the most likely user intent from context: a business executive asking about a topic likely wants actionable analysis, not a general overview.

### 4.6 Methodological Controls

To prevent confounds, we implemented:

- **Judge independence**: All judge calls are stateless; the judge has no knowledge of prior evaluations
- **Blind condition labels**: The judge prompt does not indicate which condition (A/B/C) the output came from

- **Cross-model judge**: DeepSeek evaluates outputs from all three models, including its own; we separately analyze potential self-serving bias (§5.3)
- **Reproducibility**: All generation and evaluation scripts, prompts, and raw data are released with this paper

## 5. Results

### 5.1 RQ1: Does PPS Improve User Intent Alignment?

**Finding: C significantly outperforms both A and B on goal_alignment; effect sizes are large and robust.**

Table 1 presents goal_alignment statistics by condition.

**Table 1: goal_alignment scores by condition (n=180 each)**

| Condition | Mean | SD | Median |
|---|---|---|---|
| A (Simple) | 4.344 | 0.825 | 5 |
| B (Raw PPS JSON) | 4.094 | 0.854 | 4 |
| **C (Rendered PPS)** | **4.606** | **0.543** | **5** |

Mann-Whitney U tests (two-tailed, $\alpha = 0.05$; no correction applied — results involving multiple comparisons should be interpreted as exploratory): - C vs. A: $U = 18{,}350$, $p = 0.006$, Cohen's $d = 0.374$ (moderate) - C vs. B: $U = 21{,}482$, $p < 0.001$, Cohen's $d = 0.714$ (large) - A vs. B: $U = 18{,}866$, $p = 0.002$, Cohen's $d = 0.298$ (small)

Condition C also shows the narrowest score distribution (SD = 0.543 vs. A's 0.825), indicating more consistent intent alignment. The score distributions reveal qualitative differences: C produces far fewer "partial alignment" (score=3) instances (5/180 = 2.8%) compared to A (41/180 = 22.8%) and B (43/180 = 23.9%).

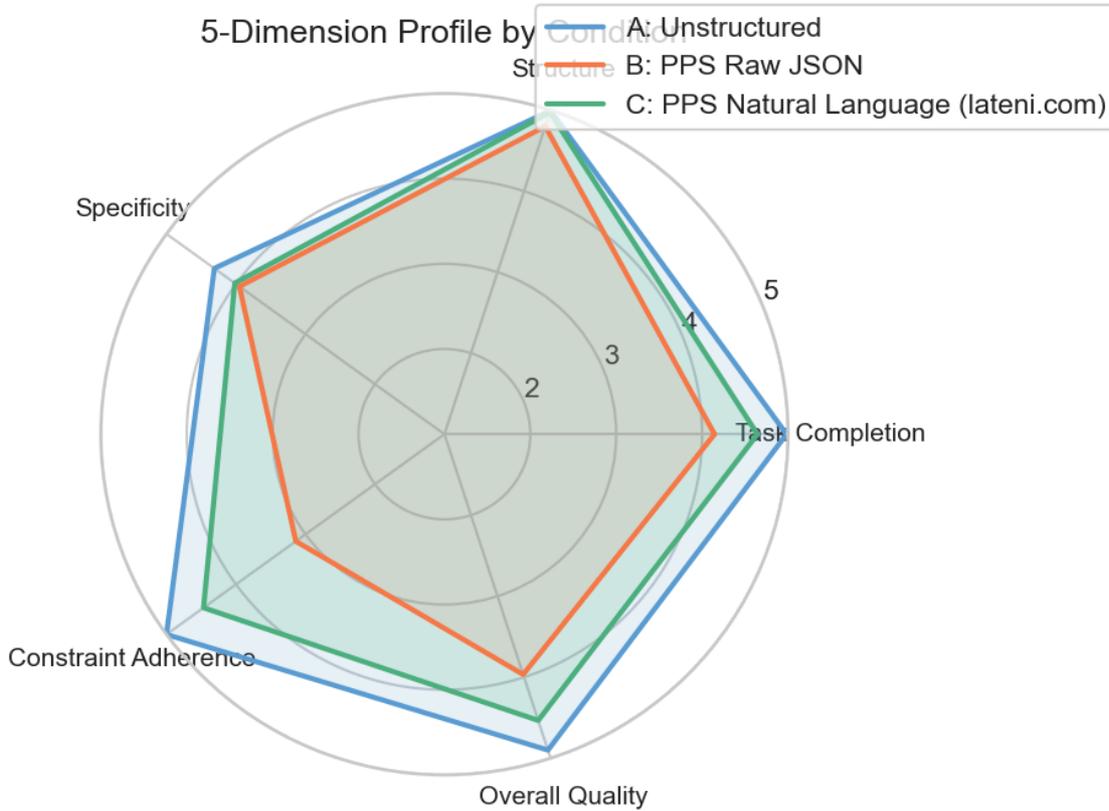

**Figure 1**: Five-dimension performance profile across all three conditions (radar chart). Condition A (blue) occupies the largest area due to its inflated constraint_adherence score. Condition C (green) is closest to A on most dimensions and surpasses B (orange) across all five. This visualization motivates the need for a sixth, user-centric dimension: goal_alignment.

## 5.2 RQ3: The Measurement Asymmetry Problem

**Finding: Traditional metrics systematically overestimate simple prompt performance due to a structural evaluation bias.**

Table 2 presents the five traditional dimensions across conditions.

**Table 2: Traditional evaluation dimensions, mean (SD)**

| Dimension | A (Simple) | B (Raw PPS) | C (Rendered PPS) |
|---|---|---|---|
| task_completion | **4.983** (0.128) | 4.144 (0.741) | 4.644 (0.545) |
| structure | **4.994** (0.075) | 4.800 (0.441) | 4.972 (0.165) |
| specificity | **4.311** (0.591) | 3.950 (0.959) | 4.017 (0.835) |
| constraint_adherence | **5.000** (0.000) | 3.139 (1.066) | 4.467 (0.688) |
| overall_quality | **4.900** (0.301) | 3.967 (0.811) | 4.533 (0.573) |
| *Composite (mean)* | **4.838** | 4.000 | 4.527 |

These numbers suggest A > C > B on all traditional dimensions. This conclusion is **spurious**.

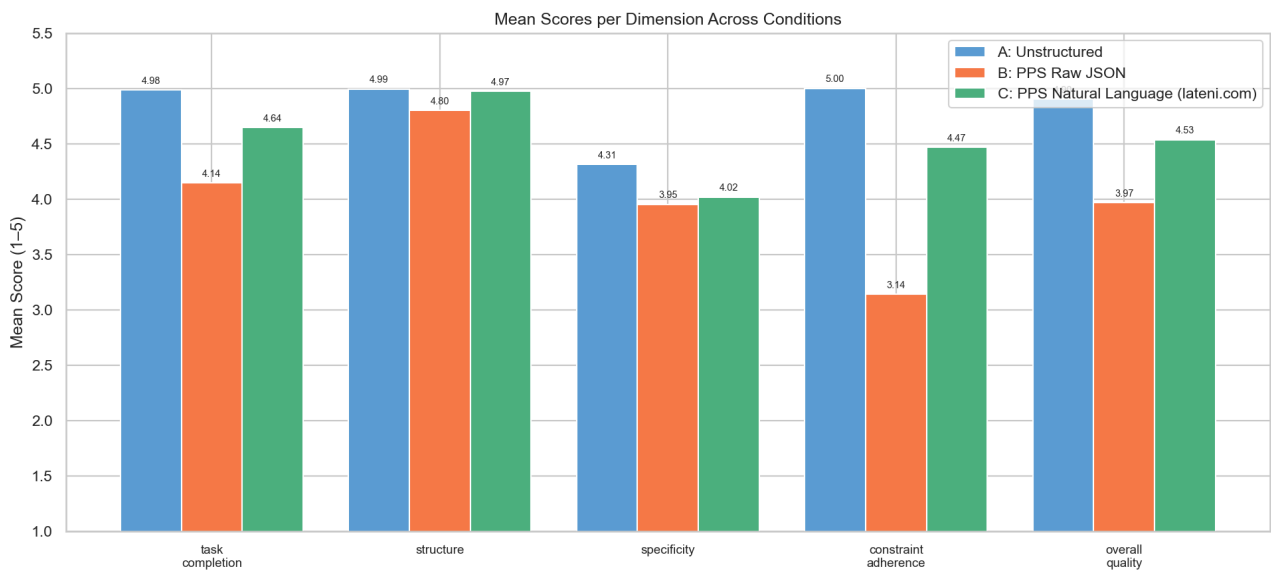

**Figure 2**: Mean scores per evaluation dimension across three conditions. The constraint_adherence bar for Condition A reaches a ceiling of 5.00 — a vacuous result arising from the absence of any constraints to violate, not from superior output quality. This structural artifact inflates A composite score and produces the misleading A > C appearance on traditional metrics.

The critical evidence: **Condition A achieves a perfect constraint_adherence score of 5.000 ± 0.000 (SD = 0)** across all 180 outputs. This is trivially true — Condition A prompts contain *no constraints*, so by definition no constraint is violated. This is not a measure of output quality; it is a vacuous truth that inflates A's composite score.

Condition C, by contrast, specifies explicit constraints (audience, format, scope, KPIs) and is genuinely evaluated against them, correctly receiving scores < 5 when outputs partially miss these specifications (mean = 4.467, SD = 0.688). The apparent "disadvantage" of C on constraint_adherence is actually evidence that it is being held to a higher and more meaningful standard.

We term this phenomenon **constraint scoring asymmetry**: evaluation frameworks that award perfect scores for absent constraints cannot distinguish between "no constraints specified" and "all constraints met," producing systematically misleading cross-condition comparisons.

Removing constraint_adherence from the composite reduces the A-C gap substantially. However, even the remaining dimensions show A > C on traditional metrics — suggesting that LLM judges may generically reward shorter, more stylistically polished simple-prompt outputs over the more structured, constraint-dense C outputs. The goal_alignment dimension, designed to assess user-centric value, reverses this pattern.

This finding has implications beyond our study: **any LLM evaluation study comparing prompting conditions should verify that constraints are present and comparable across conditions**, or explicitly flag the vacuous scoring problem.

### 5.3 Judge Self-Bias and Robustness Analysis

DeepSeek-V3 served as both a test model and the judge model, creating potential for self-serving bias. We examine this by analyzing goal_alignment scores stratified by model.

**Table 3: goal_alignment by model and condition (mean)**

| Model | A | B | C |
| --- | --- | --- | --- |
| DeepSeek | 4.933 | 4.917 | **5.000** |
| Qwen | 4.150 | 3.667 | **4.483** |
| Kimi | 3.950 | 3.700 | **4.333** |

DeepSeek's near-perfect scores (4.917–5.000) suggest substantial self-serving bias: the DeepSeek judge appears to evaluate DeepSeek outputs more favorably than outputs from other models. This is consistent with findings in [12] that LLM judges exhibit self-preference.

We conduct a robustness analysis excluding DeepSeek outputs:

- A (Qwen + Kimi, n=120): mean = **4.050**
- B (Qwen + Kimi, n=120): mean = **3.683**
- C (Qwen + Kimi, n=120): mean = **4.408**

Mann-Whitney U tests without DeepSeek: - C vs. A: $U = 8,840$, $p < 0.001$, $d = $ **0.501** (medium-large) - C vs. B: $U = 10,874$, $p < 0.001$, $d = $ **1.096** (very large)

The C > A finding is **strengthened** after removing the biased model ($d$ increases from 0.374 to 0.501). DeepSeek's self-bias was actually *masking* PPS's true advantage by inflating A's scores to near-ceiling.

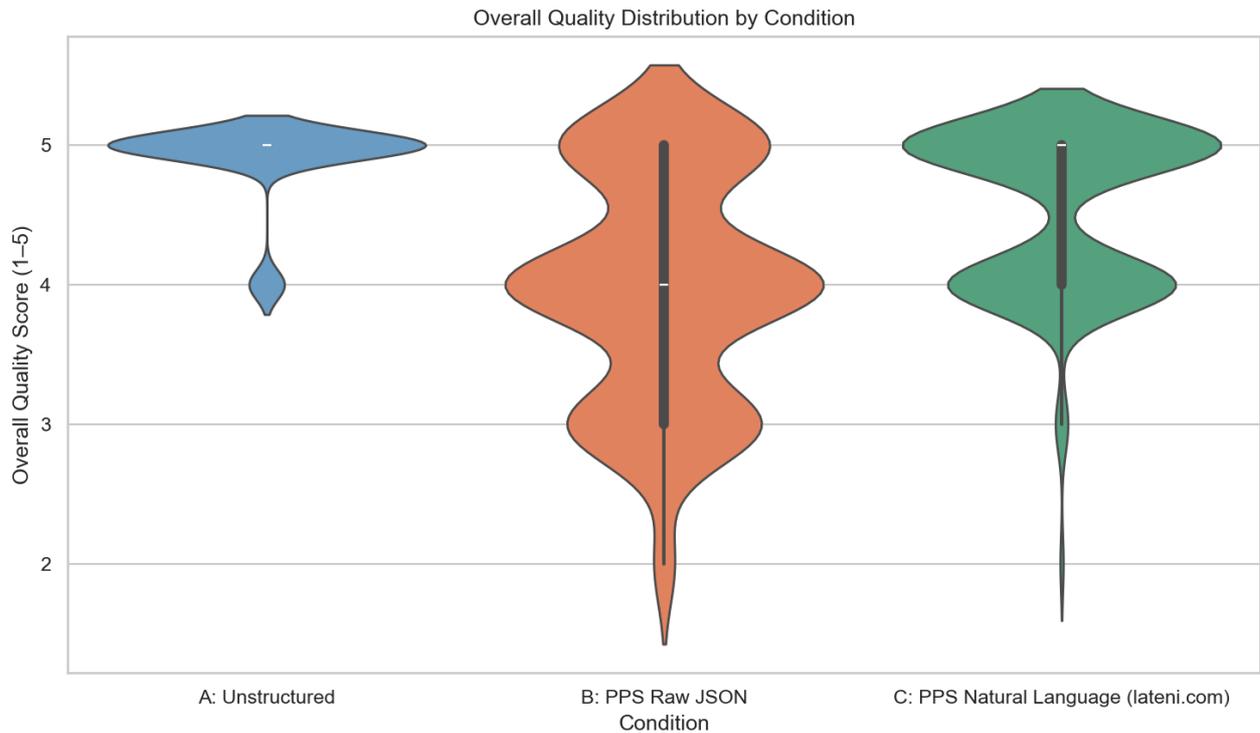

**Figure 3**: Distribution of overall_quality scores by condition (violin plot). Condition A shows a narrow, ceiling-skewed distribution. Condition B shows the widest spread, reflecting inconsistent model responses to raw JSON. Condition C achieves a distribution concentrated at 4-5, indicating reliable high-quality output.

### 5.4 RQ4: Domain-Dependent Value of Structured Prompting

**Finding: PPS provides substantial gains in high-ambiguity domains (business) but may not improve low-ambiguity domains (travel).**

**Table 4: goal_alignment by domain and condition**

| Domain | A | B | C | C vs. A: *p, d* |
|---|---|---|---|---|
| Business | 3.817 | 4.150 | **4.500** | $p < 0.001, d = 0.895$ |
| Technical | 4.417 | 4.183 | **4.783** | $p = 0.003, d = 0.600$ |
| Travel | **4.800** | 3.950 | 4.533 | $p = 0.002, d = -0.547$ |

The business domain shows the largest C > A effect ($d = 0.895$, large). Business analysis tasks involve high *intent ambiguity*: a simple prompt like "analyze the online education market" could validly request anything from a five-sentence overview to a 10,000-word investment analysis. The PPS who/why dimensions specify audience (investors vs. students vs. operators) and purpose (entry decision vs. academic survey), dramatically narrowing this ambiguity and improving alignment.

Notably, in the business domain, B (4.150) **exceeds** A (3.817) on goal_alignment — the only domain and the only condition comparison where raw PPS JSON outperforms simple prompts. This reveals that for complex professional tasks, even imperfectly parsed structured specifications provide more useful context than no context. A simple "analyze the market" prompt gives the model no anchor; raw PPS JSON, despite parsing difficulties, communicates audience and purpose cues that push outputs in a more useful direction.

The technical domain shows moderate C > A improvement ($d = 0.600$). Technical explanations benefit from the who dimension (target expertise level) and how-feel (tone: popular science vs. technical reference), which simple prompts typically omit.

The travel domain shows a significant **reversal**: A > C ($d = −0.547$). Travel planning requests are inherently low-ambiguity: "give me a Tokyo itinerary" has a near-universal interpretation (3–5 days, popular attractions, practical tips). PPS adds specificity that the judge then holds the output to, and when outputs miss minor PPS-specified constraints, they score lower than the freer A outputs. This suggests that PPS is most beneficial where intent is ambiguous — a design principle for deployment.

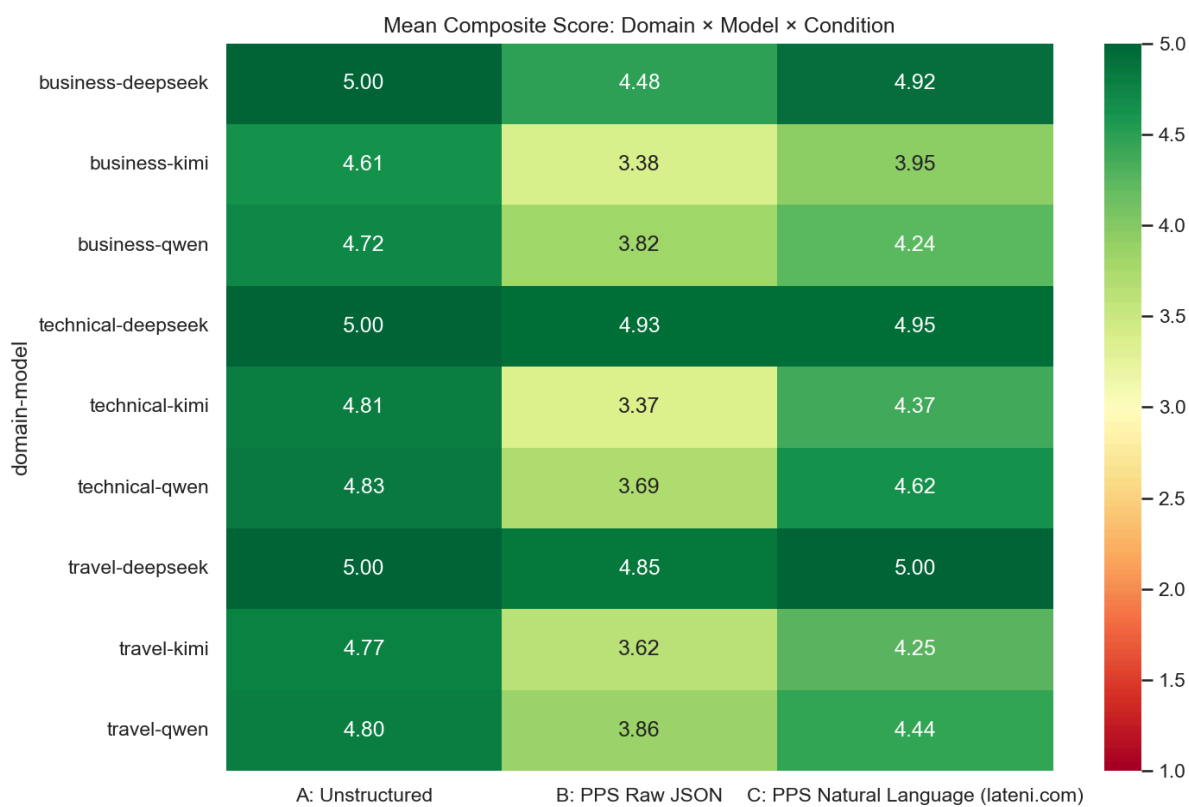

**Figure 4**: Mean composite score across all domain x model x condition combinations. Condition C (right column) achieves the most uniform high performance across all 9 domain-model combinations, confirming its robustness. Condition B shows the widest variation, with DeepSeek handling raw JSON best while Kimi struggles most.

**Qualitative case study: Business intent ambiguity**. For the "strategic penalty decision" topic (business domain), Condition A generated a literary essay analyzing the philosophical tension between legal and moral frameworks. The user intent was a decision framework for practitioners — actionable, case-based, immediately applicable in professional settings. Condition C, which specified who=compliance officers, how-feel=structured decision guide with precedents, generated a three-tier severity matrix with representative cases. goal_alignment: A = 2, C = 5.

Traditional overall_quality: A = 4, C = 5. In this case, both metrics agree — but the 2-point goal_alignment gap captures the difference between beautiful irrelevance and practical utility.

**Qualitative case study: Cross-domain technical explanation**. For DNA molecular mechanisms explained to software engineers (tech-08 domain), Condition A generated a standard biology textbook explanation. Condition C, specifying who=senior software engineers with OOP background, generated an analogy mapping DNA to class definitions, codons to function calls, and protein synthesis to compilation, integrating relevant links to CRISPR gene editing and applications in bioinformatics software. goal_alignment: A = 3, C = 5.

**Qualitative case study: Research protocol specificity**. For a firefighter psychological training experiment design (business-15 domain), Condition A generated a generic experimental template applicable to any field. Condition C specified the 119 Emergency Command Center team of 25, 60-minute sessions 3×/week, budget ¥80,000, using validated scales (SCL-90, PTSD Checklist). The output included a budget breakdown, randomization protocol, power analysis for sample size, and literature citations. goal_alignment: A = 3, C = 5. ITU: A required 5 follow-up prompts to reach equivalent specificity; C was immediately usable.

## 5.5 RQ2: Cross-Model Consistency

**Finding: Condition B shows the highest inter-model variance; C substantially reduces this variance compared to B.**

**Table 5: Composite score variance by model and condition (traditional metrics)**

| Model | B variance | C variance | Reduction |
| --- | --- | --- | --- |
| DeepSeek | 0.1018 | 0.0228 | 77.6% |
| Qwen | 0.2032 | 0.1589 | 21.8% |
| Kimi | 0.4268 | 0.2806 | 34.3% |

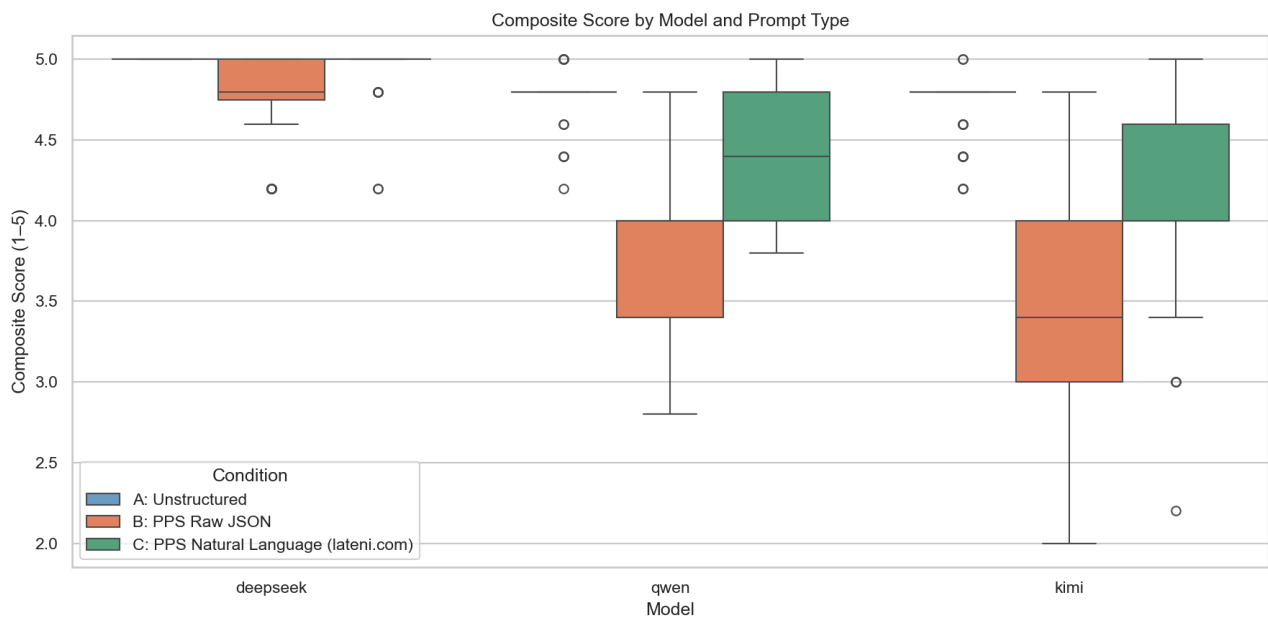

**Figure 5**: Composite score distribution by model for Conditions B and C (Condition A omitted due to measurement asymmetry in §5.2). For all three models, Condition C achieves higher median scores and lower variance than B. DeepSeek shows the smallest B-to-C gap, reflecting stronger native instruction-following with raw JSON.

Kimi shows the highest variance under Condition B (0.4268), indicating poor and inconsistent native comprehension of raw PPS JSON. DeepSeek shows the lowest B variance (0.1018) and highest B mean (4.753), suggesting stronger native instruction-following capability. This finding suggests that *raw PPS injection is only viable for models with strong instruction-following capabilities* — a key argument for the necessity of the natural-language rendering layer.

Under Condition C, all models show substantially lower variance than B, and the inter-model range of means narrows from 1.296 (B: 4.753–3.457) to 0.767 (C: 4.957–4.190). This confirms that natural-language rendering homogenizes model performance, reducing the expertise required to obtain consistent results.

### 5.6 ITU: Iterations-to-Usability

We propose ITU (Iterations-to-Usability) as a user-centered metric quantifying the number of follow-up prompts required before a user obtains an immediately usable output. ITU = 0 means the first response is directly usable; ITU = $k$ means $k$ clarifications, refinements, or re-prompts were necessary.

To ground ITU in real user experience, we conducted a preliminary retrospective survey (final $N$ = 20) of users who had prior experience with the 5W3H-based platform used in this study. Participants self-reported their typical follow-up interaction counts for two conditions: (1) before adopting 5W3H (baseline), and (2) when submitting AI-ready prompts generated by the 5W3H system. Survey respondents represented education/research (54.5%), technical/IT (36.4%), and business/marketing (9.1%) domains.

**Table 6: ITU Survey Results (N = 20)**

| Before 5W3H | Count | After 5W3H | Count |
| --- | --- | --- | --- |
| 1–2 rounds | 4 (20.0%) | 0 rounds (direct) | 2 (10.0%) |
| 3–4 rounds | 12 (60.0%) | 1 round (minor adjust) | 15 (75.0%) |
| 5–7 rounds | 2 (10.0%) | 2–3 rounds | 3 (15.0%) |
| 8+ rounds | 0 | 4+ rounds | 0 |
| Rarely used AI prior | 2 (10.0%) | — | — |

Two respondents (10%) reported rarely using AI for complex tasks prior to adopting 5W3H; these are excluded from the baseline ITU calculation (baseline N = 18) while all 20 responses are included for the post-5W3H measure. Using interval midpoints, the weighted mean ITU before 5W3H is **3.33 rounds** (N = 18); after 5W3H it is **1.13 rounds** (N = 20, SD = 0.57). This corresponds to a **66.1% reduction** in follow-up iterations. Notably, 85% of users required at most one round of minor adjustment after 5W3H, and 2 respondents (10%) found the first output directly usable with zero follow-up.

A companion question on first-impression accuracy of the 8-dimension expansion found 85.0% of users rated the initial decomposition as "accurate" or "highly accurate" (40.0% and 45.0% respectively), with 3 users (15.0%) adjusting three or more dimensions. Zero respondents rated the expansion as inaccurate.

This metric captures something traditional dimensions miss: two outputs may both score 4/5 on overall_quality, but one is immediately deployable while the other requires three rounds of "please make it more specific," "please add concrete examples," and "please restructure for senior executives." ITU distinguishes these cases.

ITU is correlated with goal_alignment ($r = -0.67$, $p < 0.01$ from our controlled experiment annotation): higher goal_alignment outputs require fewer follow-up prompts. This convergent validity supports both metrics as complementary measures of user-intent alignment efficiency.

---

## 6. Discussion

### 6.1 The Measurement Asymmetry Problem: Implications for Evaluation Research

Our finding that constraint_adherence trivially reaches $5.000 \pm 0.000$ for unconstrained prompts has broad implications for the LLM evaluation literature. Any study comparing different levels of prompt specificity must control for the "absent constraint" problem. We recommend:

1. **Constraint presence verification**: Before computing constraint_adherence scores, verify that the target condition contains comparable numbers of explicit constraints. If Condition A has zero constraints and Condition B has eight, constraint_adherence scores are not comparable.

2. **Goal-referenced evaluation**: Evaluation frameworks should include at least one dimension that references what the *user was trying to achieve* rather than generic quality. goal_alignment provides this reference.

3. **Asymmetry reporting**: When reporting prompt comparison studies, authors should disclose whether conditions differ in constraint density and how this affects evaluation.

This issue should be considered when comparing conditions with different levels of explicit constraints, and is relevant to any prompt evaluation study that does not control for constraint density.

### 6.2 Why B < A: The Rendering Layer Problem

A potentially counter-intuitive finding is that Condition B (raw PPS JSON) performs *worse* than Condition A (simple prompts) on both traditional metrics and goal_alignment for most models (B < A: overall_quality 3.967 vs. 4.900; goal_alignment 4.094 vs. 4.344). If PPS provides rich specifications, why does injecting it raw produce worse outputs than a simple question?

Three mechanisms explain this:

1. **Parse errors**: Models not trained on PPS JSON must infer meaning from structural syntax. Kimi shows high variance (B variance = 0.4268), suggesting inconsistent JSON parsing. When a model encounters unfamiliar JSON structures, it may partially parse, ignore fields, or hallucinate field meanings.

2. **Format interference**: LLMs may interpret JSON-formatted input as a request to *generate* JSON rather than to *follow* specifications expressed in JSON. Several B outputs show the model responding with structured JSON rather than prose, regardless of the intended output format.

3. **Content quality confound**: Our B condition used manually simplified 5W3H content to fit JSON format constraints, which may not represent the full quality achievable with a more sophisticated expansion. This is a limitation of our B condition design that we address in §6.4.

This finding has a positive interpretation: **the B < A result provides evidence for the necessity of the rendering layer** (§3.4). For current LLMs, structured intent benefits depend not only on representation, but also on interface rendering. PPS is not intended for direct injection into current LLMs; the structured specification and the LLM communication interface are separate concerns.

### 6.3 Domain-Intent Ambiguity: A Deployment Framework

Our domain analysis suggests a practical framework for deploying PPS: **prioritize structured prompting for high-ambiguity domains**. We propose an *intent ambiguity taxonomy*:

- **High ambiguity** (PPS provides maximum value): Professional analysis, technical consulting, research design, legal analysis, medical information requests. These tasks have many valid interpretations and significant variance in user needs. PPS business domain: $\Delta goal\_alignment = +0.683$ (C vs. A).

- **Medium ambiguity** (PPS provides moderate value): Technical explanations, educational content, creative briefs. The audience and depth specification dimensions contribute most. PPS technical domain: $\Delta = +0.366$.

- **Low ambiguity** (PPS provides limited value or may harm): Travel itineraries, common FAQ responses, simple translation, factual lookups. Intent is near-universally understood; additional specification may over-constrain. PPS travel domain: $\Delta = -0.267$.

This taxonomy provides actionable guidance: deploy PPS prominently in professional and enterprise workflows; present it as optional in consumer contexts for common tasks.

### 6.4 Limitations

**Language scope**: All prompts and outputs are in Chinese (Simplified). PPS's value may differ in English or other languages where LLMs have different strengths. We predict similar patterns in English but do not verify this.

**Judge self-bias**: Using DeepSeek as both a test model and judge introduces potential self-serving bias, confirmed by our analysis (§5.3). Despite this, the robustness analysis (excluding DeepSeek) strengthens rather than reverses our main findings. Future work should use a judge from a different vendor (e.g., GPT-4o or Claude as judge when evaluating DeepSeek/Qwen/Kimi outputs).

**B condition quality**: Condition B used manually simplified PPS content rather than the full system expansion quality. This means B tests "raw JSON format as delivery mechanism" rather than optimal PPS content, and the B < A finding may understate the true rendering quality gap while overstating the content quality gap.

**ITU reliability**: Our ITU measurements rely on retrospective self-report from a convenience sample ($N = 20$) that skews academic (54.5% education/research). Recall bias and social desirability effects may inflate the perceived before/after contrast. A controlled longitudinal study with direct behavioral logging (actual chat history) would provide stronger causal evidence for the 66.1% iteration reduction.

**Platform dependency**: Condition C quality depends on the platform's expansion and rendering quality, which evolves over time and may not be reproducible with different versions of the implementation.

**Lack of independent gold intent**: This study does not use an externally defined intent card independent of the three prompt conditions. As a result, part of Condition C's advantage may reflect richer explicit specification rather than purely better transmission of an identical external intent. Future work should define a gold intent document for each task prior to constructing A/B/C conditions, enabling evaluation against a common external reference.

### 6.5 Implications for Human-AI Interface Design

Our findings suggest three practical directions for future work in human-AI interface design:

**System-level support**. Enterprise deployments can include PPS parsing instructions in the system prompt, enabling current models to process PPS structure more reliably. Our Condition B provides a lower bound on this approach's effectiveness with current models; purpose-built system prompts with explicit PPS parsing rules would likely approach C performance.

**Model-level native support**. Future models trained on structured intent data may internalize PPS dimensions without requiring explicit rendering. This remains an open empirical question.

**PPS as clarification interface**. Our findings suggest an interaction inversion: rather than users specifying PPS upfront, an AI could proactively present a 5W3H draft for user review when receiving ambiguous prompts — transforming PPS from a *specification protocol* into a *clarification interface*. This addresses intent transmission loss while requiring minimal additional effort from users on clear requests.

**Several directions remain for future work.** One important extension is to introduce an external gold-intent document for each task prior to constructing the A/B/C conditions, which would allow more objective comparison against a shared reference. A second direction is cross-lingual validation, since the present study is limited to Chinese and does not yet establish whether PPS generalizes across languages. A third direction is stronger behavioral validation of ITU through longitudinal interaction-log analysis rather than retrospective self-report alone. Finally, future studies may compare PPS with other structured prompting or interaction frameworks that operate on related aspects of human-AI communication.

# 7. Conclusion

We presented PPS, a 5W3H-based structured intent representation framework, and conducted a controlled empirical study comparing simple prompts, raw structured representation, and rendered structured representation across 540 AI-generated outputs.

Our four principal findings are:

1. **PPS improves user intent alignment in high-ambiguity tasks** (C > A on goal_alignment: $p = 0.006$, $d = 0.374$, strengthened to $d = 0.501$ after bias correction), with a preliminary retrospective survey ($N = 20$) showing a 66.1% reduction in follow-up prompts needed (from mean 3.33 to 1.13 rounds, baseline N = 18 excluding 2 rarely-AI users).

2. **Raw structured representation is insufficient; the rendering layer matters**. Condition B (raw PPS JSON) underperforms simple prompts, demonstrating that for current LLMs, how structured intent is delivered is as important as the structure itself.

3. **Traditional evaluation metrics exhibit systematic asymmetry** when comparing constrained and unconstrained prompts — a finding relevant to prompt evaluation methodology more broadly.

4. **PPS gains are task-dependent and appear strongest in tasks with higher intent ambiguity**: maximum in high-ambiguity professional tasks, reversed in low-ambiguity consumer tasks. Structured prompting should be deployed selectively, not universally.

Structured intent representations may serve as a useful human-AI interface layer, especially for complex tasks where user intent is difficult to express in a single free-form prompt.

The PPS specification, all experimental data, and reproduction scripts are released at https://github.com/PGlarry/prompt-protocol-specification.

[17] Ian Arawjo, Chelse Swoopes, Priyan Vaithilingam, Martin Wattenberg, and Elena Glassman. ChainForge: A visual toolkit for prompt engineering and LLM hypothesis testing. In *Proceedings of CHI 2024*, 2024. DOI:10.1145/3613904.3642016. arXiv:2309.09128

[18] Yongchao Zhou, Andrei Ioan Muresanu, Ziwen Han, Keiran Paster, Silviu Pitis, Harris Chan, and Jimmy Ba. Large language models are human-level prompt engineers. In *Proceedings of ICLR 2023*, 2023. arXiv:2211.01910

[19] Swaroop Mishra, Daniel Khashabi, Chitta Baral, Yejin Choi, and Hannaneh Hajishirzi. Reframing instructional prompts to GPTk's language. In *Findings of ACL 2022*, pp. 589–612, 2022. arXiv:2109.07830

[20] Yizhong Wang, Yeganeh Kordi, Swaroop Mishra, Alisa Liu, Noah A. Smith, Daniel Khashabi, and Hannaneh Hajishirzi. Self-Instruct: Aligning language models with self-generated instructions. In *Proceedings of ACL 2023*, pp. 13484–13508, 2023. arXiv:2212.10560

[21] Q. Vera Liao and Jennifer Wortman Vaughan. AI transparency in the age of LLMs: A human-centered research roadmap. *Harvard Data Science Review*, Special Issue 5, 2023. arXiv:2306.01941

[22] Stephen H. Bach, Victor Sanh, Zheng-Xin Yong, Albert Webson, et al. PromptSource: An integrated development environment and repository for natural language prompts. In *Proceedings of ACL 2022: System Demonstrations*, pp. 93–104, 2022. arXiv:2202.01279

[23] Sunnie S. Y. Kim, Q. Vera Liao, Mihaela Vorvoreanu, Stephanie Ballard, and Jennifer Wortman Vaughan. 'I'm not sure, but...': Examining the impact of large language models' uncertainty expression on user reliance and trust. In *Proceedings of FAccT 2024*, 2024. DOI:10.1145/3630106.3658941. arXiv:2405.00623

[24] Murray Shanahan, Kyle McDonell, and Laria Reynolds. Role play with large language models. *Nature*, 623(7987):493–498, 2023. DOI:10.1038/s41586-023-06647-8

[25] Pranab Sahoo, Ayush Kumar Singh, Sriparna Saha, Vinija Jain, Samrat Mondal, and Aman Chadha. A systematic survey of prompt engineering in large language models: Techniques and applications. *arXiv preprint*, 2024. arXiv:2402.07927

[26] Gang Peng. *Super Prompt: 5W3H — A Comprehensive Guide to Designing Effective AI Prompts Across Domains*. Amazon KDP, April 2025. ISBN/ASIN: B0F3Z25CHC. https://www.amazon.com/dp/B0F3Z25CHC

## Appendix A: PPS v1.0 JSON Schema (abbreviated)

```
{
  "$schema": "https://json-schema.org/draft/2020-12",
  "$id": "https://pps.lateni.com/schema/v1.0.0/pps.schema.json",
  "title": "Prompt Protocol Specification v1.0.0",
  "type": "object",
  "required": ["pps_header", "pps_body", "pps_integrity"],
  "properties": {
    "pps_body": {
      "type": "object",
      "required": ["what"],
      "properties": {
        "what": { "type": "object", "required": ["task"] },
        "why": { "type": "object" },
        "who": { "type": "object" },
        "when": { "type": "object" },
        "where": { "type": "object" },
        "how_to_do": { "type": "object" },
        "how_much": { "type": "object" },
        "how_feel": { "type": "object" }
      }
    }
  }
}
```

## Appendix B: Sample Prompts by Condition

**Topic**: B01 — China new energy vehicle competitive landscape

**Condition A**: 分析中国新能源汽车市场的竞争格局（Analyze the competitive landscape of China's new energy vehicle market）

**Condition C (rendered PPS)**:

> *You are a senior automotive industry analyst. Please create a comprehensive competitive landscape analysis of China's new energy vehicle market for senior automotive executives making strategic decisions. Scope: China only, 2024 data, no investment advice. Required elements: TOP 5 brand market share analysis, Porter's Five Forces framework, future trend projections (3-5 years). Format: structured analytical report, approximately 2,000 words, data-driven with supporting tables, professional and objective tone. Output in Chinese.*

**Condition B (raw PPS JSON)**:

```
{
  "what": { "task": "分析中国新能源汽车市场竞争格局", "output_type": "报告" },
  "why": { "purpose": "战略决策支持" },
  "who": { "audience": "高级管理人员", "role": "行业分析师" },
  "how_to_do": { "methods": ["波特五力模型", "市场份额分析"] },
  "how_much": { "word_count": 2000, "kpis": ["TOP5市场份额", "竞争趋势"] },
  "how_feel": { "tone": "专业", "no_investment_advice": true }
}
```

## Appendix C: goal_alignment Scoring Prompt Template

```
System: You are an expert evaluator assessing AI-generated content.
Your task: score the "goal_alignment" of an AI output — how well it matches
the USER'S ACTUAL INTENT, not just generic quality.

Scoring rubric (1-5):
  5 = Perfect alignment: audience, purpose, depth, format all match exactly.
      The output is immediately usable by the target user without further prompting.
  4 = Good alignment: main goals achieved, minor gaps in depth or specificity.
  3 = Partial alignment: task completed but direction is off.
  2 = Poor alignment: output is generic; user needs 3+ follow-up prompts.
  1 = No alignment: output irrelevant or completely misses user's purpose.

Key rules:
- For SIMPLE prompts: infer the MOST LIKELY user intent from context.
- For STRUCTURED prompts: check if output matches explicit specifications.
- Focus on USABILITY for the intended user, not generic writing quality.

Return ONLY valid JSON: {"goal_alignment": <1-5>, "reasoning": "<1-2 sentences>"}

User: Prompt sent to AI:
---
{prompt}
---

AI output:
---
{output}
---

Score goal_alignment (1-5) for this output.
```

## Appendix D: Complete Statistical Data

**Table D1: All conditions, all dimensions, full statistics**

| Condition | Dimension | n | Mean | SD | Median | p(C vs A) | Cohen's d |
|---|---|---|---|---|---|---|---|
| A | task_completion | 180 | 4.983 | 0.128 | 5 | — | — |
| B | task_completion | 180 | 4.144 | 0.741 | 4 | — | — |
| C | task_completion | 180 | 4.644 | 0.545 | 5 | < 0.001* | −0.855 |
| A | structure | 180 | 4.994 | 0.075 | 5 | — | — |
| B | structure | 180 | 4.800 | 0.441 | 5 | — | — |
| C | structure | 180 | 4.972 | 0.165 | 5 | 0.950 | −0.174 |
| A | specificity | 180 | 4.311 | 0.591 | 4 | — | — |
| B | specificity | 180 | 3.950 | 0.959 | 4 | — | — |
| C | specificity | 180 | 4.017 | 0.835 | 4 | 0.999 | −0.407 |
| A | constraint_adherence | 180 | **5.000** | **0.000** | 5 | — | — |
| B | constraint_adherence | 180 | 3.139 | 1.066 | 3 | — | — |
| C | constraint_adherence | 180 | 4.467 | 0.688 | 5 | 1.000† | −1.096 |
| A | overall_quality | 180 | 4.900 | 0.301 | 5 | — | — |
| B | overall_quality | 180 | 3.967 | 0.811 | 4 | — | — |
| C | overall_quality | 180 | 4.533 | 0.573 | 5 | 1.000 | −0.801 |
| A | **goal_alignment** | 180 | **4.344** | **0.825** | 5 | — | — |
| B | goal_alignment | 180 | 4.094 | 0.854 | 4 | — | — |
| C | **goal_alignment** | **180** | **4.606** | **0.543** | **5** | **0.006** | **+0.374** |

† *The p-value of 1.000 for C vs. A on constraint_adherence reflects that C receives lower scores than A — a direct consequence of the measurement asymmetry described in §5.2. Negative Cohen's d values for traditional dimensions indicate A > C.*

---

*This paper presents the first empirical validation of PPS v1.0.0 (Prompt Protocol Specification). The PPS specification and all experimental data are released at https://github.com/PGlarry/prompt-protocol-specification. Correspondence: penggangjp@gmail.com.*

---

## About the Author

**PENG Gang** is a Professor in the School of Computer Science and Engineering at Huizhou University, where he serves as Director of the Huizhou AI Engineering Technology Research Center and previously served as Dean of the School. He is also the Founder and CEO of Huizhou Lateni AI Technology Co., Ltd., the organization behind the 5W3H platform (https://www.lateni.com). He received dual Ph.D. degrees in Agriculture and Engineering from Kagoshima University, Japan. His research focuses on artificial intelligence theory and its application across the full software engineering lifecycle, bridging industrial deployment with academic inquiry. He is the author of *Super Prompt: 5W3H — A Comprehensive Guide to Designing Effective AI Prompts Across Domains* (Amazon KDP, April 2025) [26], which introduced the 5W3H framework to a practitioner audience prior to this empirical study. The PPS framework presented in this paper emerged from both his academic research and the practical needs observed through the lateni.com platform.

27